\documentclass[sigconf]{acmart}
\usepackage{multirow}
\usepackage{graphicx}
\usepackage{multicol} 
\usepackage{balance}  

\usepackage{caption}
\usepackage{subcaption}
\usepackage{soul}

\AtBeginDocument{%
  }

\setcopyright{acmcopyright}
\copyrightyear{2023}
\acmYear{2023}
\acmDOI{XXXXXXX.XXXXXXX}

\acmConference[Conference acronym 'XX]{Make sure to enter the correct
  conference title from your rights confirmation emai}{August 6-10, 2023}{Long Beach, CA}
\acmPrice{15.00}
\acmISBN{978-1-4503-XXXX-X/18/06}

\begin{document}


\title{Novel Physics-Based Machine-Learning Models for Indoor Air Quality Approximations}

\author{Ahmad Mohammadshirazi}
\email{mohammadshirazi.2@osu.edu}
\affiliation{%
  \institution{The Ohio State University}
  \city{Columbus}
  \state{Ohio}
  \country{USA}
}

\author{Aida Nadafian}
\email{aidanada@bu.edu}
\affiliation{%
  \institution{Boston University}
  \city{Boston}
  \state{Massachusetts}
  \country{USA}
}

\author{Amin Karimi Monsefi}
\email{karimimonsefi.1@osu.edu}
\affiliation{%
  \institution{The Ohio State University}
  \city{Columbus}
  \state{Ohio}
  \country{USA}
}

\author{Mohammad H. Rafiei}
\email{mrafiei1@jhu.edu}
\affiliation{%
  \institution{Johns Hopkins University}
  \city{Baltimore}
  \state{Maryland}
  \country{USA}
  }

\author{Rajiv Ramnath}
\email{ramnath.6@osu.edu}
\affiliation{%
  \institution{The Ohio State University}
  \city{Columbus}
  \state{Ohio}
  \country{USA}
}

\renewcommand{\shortauthors}{Mohammadshirazi et al.}

\begin{abstract}

Cost-effective sensors are capable of real-time capturing a variety of air quality-related modalities from different pollutant concentrations to indoor/outdoor humidity and temperature. Machine learning (ML) models are capable of performing air-quality "ahead-of-time" approximations. Undoubtedly, accurate indoor air quality approximation significantly helps provide a healthy indoor environment, optimize associated energy consumption, and offer human comfort. However, it is crucial to design an ML architecture to capture the domain knowledge, so-called problem physics. In this study, we propose six novel physics-based ML models for accurate indoor pollutant concentration approximations. The proposed models include an adroit combination of state-space concepts in physics, Gated Recurrent Units, and Decomposition techniques. The proposed models were illustrated using data collected from five offices in a commercial building in California. The proposed models are shown to be less complex, computationally more efficient, and more accurate than similar state-of-the-art transformer-based models. The superiority of the proposed models is due to their relatively light architecture (computational efficiency) and, more importantly, their ability to capture the underlying highly nonlinear patterns embedded in the often contaminated sensor-collected indoor air quality temporal data. 

\end{abstract}



\keywords{Time Series, State-Space, Recurrent Neural Networks, Gated Recurrent Units, Decomposition, Fourier, Air Quality}


\maketitle

\section{Introduction}
\label{sec:intro}

The quality of the indoor environment significantly impacts human health, performance, satisfaction, and productivity in built environments. One important aspect is indoor air quality (IAQ), which can negatively affect the comfort and health of occupants and has been linked to "sick building syndrome" \cite{mujan2019influence}. However, public awareness has been on outside air quality (OAQ) \cite{turner2020outdoor}, and past research has also paid most attention to the impact of OAQ on health and wellness rather than IAQ \cite{ghaffarianhoseini2018sick}. In fact, the air inside buildings can be worse than the air outside and have greater ill effects, especially given that people spend a significant amount of time within office buildings. 

Indoor pollutants can be divided into three categories: gases, biological contaminants, and particulate matter \cite{gonzalez2021state}. Sources of these pollutants include chemicals used indoors, such as low-emitting paints, sealants, and adhesives, as well as appliances, such as stoves and heaters. Among these pollutants, carbon dioxide ($CO_{2}$) concentration is considered the most important indicator of IAQ. High $CO_{2}$ levels can have a range of negative effects on people, including poor health and decreased cognitive performance. Studies have shown a correlation between high $CO_{2}$ levels in indoor environments and negative impacts on human health \cite{zhang2021associations, jacobson2019direct}. Addressing ($CO_{2}$) can also bring other benefits, such as energy savings, cost reduction, improved health and productivity, and an environment that promotes talent cultivation.

Buildings are a significant source of energy consumption and greenhouse gas emissions, and finding ways to reduce energy use while maintaining healthy indoor environments is a significant challenge. Both forced and natural ventilation provides a set of means for improving IAQ; however, it must be managed depending on the quality of the air outside. A second way is to use air cleaning technologies such as air purifiers and air filters. These devices can help to remove particles, allergens, and other pollutants from the air. 

Finally, employing sensors to monitor airborne pollutants can help to optimize building energy use and maintain healthy IAQ. For example, sensors could monitor IAQ and adjust ventilation rates to bring in more fresh outdoor air when indoor pollution levels are high. Besides, based on sensor data, certain appliances or systems could be automatically turned off when they are not needed. In addition to reducing energy consumption and greenhouse gas emissions, optimizing building energy use in this way can also help to improve the comfort and health of building occupants \cite{huovila2009buildings}.

While indoor air automatic sensing and controlling techniques are essential, prediction is key to successfully managing pollutants. One reason is that management techniques have lead times before they become effective. In addition, and more importantly, management needs to be nuanced based on what other factors - such as temperature and outside pollutant concentrations - are likely to be. Predictive modeling techniques have been used to forecast future concentrations of pollutants and optimize energy use accordingly \cite{grange2019using}. These techniques primarily used simple mathematical pollution models built from the underlying physics and computational simulation techniques. However, given the variety of factors from indoor/outdoor temperature, different pollutant concentrations, ventilation flow, and time, simple mathematical simulations seem incapable of accurately capturing the nonlinearity of such prediction models \cite{rafiei2016neural}. 

'Problem physics' in this study refers to the key principles of IAQ, including factors like pollutant sources and ventilation. Accordingly, our machine learning model integrates these principles to accurately predict IAQ patterns. Machine Learning (ML) models have shown significant capability in identifying and capturing nonlinear relationships hidden within any data type, whether contaminated, large-scale, multi-modal, or temporal. In particular, data retrieved from sensors are time-dependent, which makes time-elapse fluctuations, frequencies, and patterns important factors in the predictions mentioned above \cite{mishra2022analyzing}. These data suit ML models such as recurrent neural networks (RNN), where the multi-modal temporal data are the primary focus for prediction tasks \cite{medsker2001recurrent}. However, RNNs are generic models that do not provide mechanisms \cite{ruehle2017evolving}; that is especially suited for the particular problem of IAQ. We posit that developing ML model architectures that leverage the physical nature of the problem would significantly boost model performance in any prediction tasks \cite{xie2023high,rafiei2022self}. In this paper, we incorporate the concept of a "state-space" into our RNN models to bring in IAQ physics and domain knowledge. A "state-space" is a mathematical representation of the possible states of a physical system, where each point corresponds to a unique state of the system. It may be used to describe the dynamics of any system, such as in IAQ concentrations  and their change over time. In a state-space RNN, a state-space may be used to guide the design of models for more accurate IAQ approximations. 

This paper demonstrates six novel physics-based RNN models on five datasets on carbon dioxide concentrations collected by sensors from five offices in a commercial building.   These six models are (1) State Space Recurrent Neural Network (SS-RNN), (2) State Space Gated Recurrent Unit (SS-GRU), (3) Decomposition State Space Recurrent Neural Network (D-SS-RNN), (4) Decomposition State Space Gated Recurrent Unit (D-SS-GRU), (5) Fourier Space Recurrent Neural Network (FD-SS-RNN), and (6) Fourier Decomposition State Space Gated Recurrent Unit (FD-SS-GRU). Each model incorporates IAQ physics, but in different ways using a \textit{progression} of RNN architectures. For example, one set consists of models with and without gated recurrent units (GRU) and with and without decomposition. The role of a GRU is to decontaminate the input data so as to include only the information that would best help in accurate air quality predictions. Decomposition, on the other hand, is about denoising the records, removing the high-frequency fluctuations to capture the short- and long-term patterns \cite{wen2019robuststl, rafiei2018novel}. The performance of the proposed models is compared with each other and  to that of state-of-the-art non-physics-based models such as Transformers \cite{zhou2021informer, lim2021temporal}, FEDformer \cite{zerveas2021transformer}, and Informer \cite{zhou2021informer}. The contribution of this paper is a novel exploration of how physical models and ML models can be combined for predicting indoor $CO_{2}$ concentrations. 

The rest of this paper is organized into five sections. Section 2 provides paper objectives. Section 3 describes related work. Section 4 explains methodology with a focus on the six proposed physics-based models. Section 5 considers results and discussion. The paper ends with a summary and conclusions in section 6.

\section{Paper Objectives}
\label{sec:problem}

The main objective of this study is to develop accurate, robust, and computationally efficient ML approximation models that are carefully designed and prepared based on "IAQ physics" and its specific, highly nonlinear, and challenging nature. We compare various novel state-of-the-art ways of introducing IAQ physics into ML, including the concept of physical state-spaces, gated recurrent units, decomposition, and Fourier transformations. Various architectures and hyperparameters are investigated for each model. The performance of the models are evaluated using the Mean Squared Error metric and compared with three baseline models that use state-of-the-art, non-physics-based techniques.

\section{Related Work}
\label{sec:related}

IAQ models are commonly used to enhance the comfort of building occupants and/or lower indoor air pollutants. One example is a study that implemented a control model with a filter to decrease indoor $CO_{2}$ levels in a sports center, using fuzzy inference to control the $CO_{2}$ concentration\cite{omarov2020fuzzy}. Another study found that a Multilayer Perceptron (MLP) algorithm was more accurate and efficient in predicting $CO_{2}$ changes compared to ML models such as Support Vector Machine (SVM), AdaBoost (AdB), Random Forest (RF), Gradient Boosting (GB), and Logistic Regression (LR); it was able to reduce 51.4\% of energy consumption\cite{taheri2021learning}. Statistical and simple ML models have also been used for IAQ predictive modeling. In a study by Lagesse et al. \cite{lagesse2020predicting}, a comparison was made between various statistical models (e.g. multiple linear regression, partial least squares regression, distributed lag model, and most minor absolute shrinkage selector operators) and ML models (e.g. simple artificial neural networks and long-short term memory) for predicting particle concentrations in outdoor air pollution. The results showed that ML models performed better than statistical models in predicting PM2.5 concentrations. However, the study did not include any modifications to physics-based models or analysis of input variables \cite{iskhakov2021review,lagesse2020predicting}.

Identifying the variables and patterns associated with IAQ can be challenging for several reasons, including but not limited to (1) the computation time, costs, and often inaccurate measurement, (2) missing data\cite{alsaber2021handling, alkabbani2022improved}, and (3) hardware problems and human errors \cite{ rodenas2022review, zafra2023novel}. Mohammadshirazi et al. \cite{mohammadshirazi2022predicting} employed about six months of IAQ data records captured from low-cost sensors in a commercial building in California to machine-approximate IAQ variables. Their ML models include the combinatory pattern recognition of Rafiei et al. \cite{rafiei2019predicting}, and different variations of Long-Short Term Memory towers to identify the most effective combination of sensor-recorded variables approximating IAQ. Moving from simple ML to more sophisticated ones, Tzoumpas et al. \cite{tzoumpas2022data}, and Chen et al. \cite{chen2022intelligent} both utilized a ML model called CNN-BiLSTM to forecast indoor $CO_{2}$ concentration. The CNN-BiLSTM model combines the strengths of two different neural network architectures, Convolutional Neural Networks (CNNs) and Bi-Directional Long Short-Term Memory (Bi-LSTM), to make predictions. CNNs are commonly used for image and signal processing tasks, while Bi-LSTM is a type of recurrent neural network to learn from past and the future context in temporal records. By using these two architectures together, the model is able to process time-series data and make accurate predictions effectively. 

Transformers \cite{zhou2021informer, lim2021temporal}, FEDformer \cite{zerveas2021transformer}, and Informer \cite{zhou2021informer} are all state-of-the-art ML models with great potentials in the field of IAQ. They are effective in forecasting, anomaly detection, and data imputation, especially in multi-scale sequential and temporal data \cite{pham2022mst}. Transformers have shown functional modeling ability for long-range dependencies and interactions in sequential data for time series modeling; as such, they harvest the nonlinear pattern hidden in both short- and long-term temporal sequences. Many different Transformers have been proposed to address particular challenges in time series modeling, such as forecasting \cite{zhou2021informer, lim2021temporal} and classification \cite{liu2021gated, 
 monsefi2022will, zerveas2021transformer}. To capture the time series global perspective, Zhou et al. \cite{zerveas2021transformer} introduced FEDformer, which combines Transformer \cite{han2021transformer} with the seasonal-trend decomposition \cite{wen2019robuststl} method. 

 Our work adds domain-specific capability to these new types of models by incorporating the physics of the phenomena under study; that is we add domain knowledge. SRNN is inspired by the state-space concept in physics. It refers to the mathematical space that describes the possible states of a physical system. The state of a system is defined by a set of variables, called state variables, that fully describe the system at a given point in time. A state-space is the set of all possible values these state variables can take on. The dimension of a state-space is equal to the number of state variables. For example, the state-space of a simple harmonic oscillator is a two-dimensional space defined by the position and velocity of the oscillator. State-space models are also used in control systems theory, where the state-space representation of a dynamic system is a compact and convenient way to model and analyze the system's behavior \cite{elkafafy2022robust}. In this paper, we benefit from this concept to create RNNs customized to the IAQ problem for more accurate estimations.

\section{Methodology}
\label{sec:method}

In this section, we first describe the data collected from five offices in a commercial building in Berkeley, California. Next, we will describe a progression through six proposed physics-based ML models. The models are (1) State-Space Recurrent Neural Network (SS-RNN), where RNN incorporates State-Space concept in physics, (2) State-Space Gated Recurrent Unit (SS-GRU), which is the SS-RNN but with GRU units to decontaminate input data (3) Decomposition State-Space Recurrent Neural Network (D-SS-RNN) including moving-average decomposition to reduce the high-frequency noise in temporal data (4) Decomposition State-Space Gated Recurrent Unit (D-SS-GRU) where GRU and decomposition techniques are combined in SS-RNN (5) Fourier Space Recurrent Neural Network (FD-SS-RNN), where focuses on the frequency domain of the temporal records and (6) Fourier Decomposition State-Space Gated Recurrent Unit (FD-SS-GRU) similar to D-SS-GRU but in the frequency domain. The proposed models are compared with each other in addition to three state-of-the-art methods in the literature that are the Transformer model of Vaswani et al. \cite{vaswani2017attention}, the Informer model of Zhou et al. \cite{zhou2021informer}, and the FEDformer model of Zhou et al. \cite{zhou2022fedformer}; these transformer models are also briefed in this section.

\subsection{Data}

Researchers from the Lawrence Berkeley National Laboratory in California collected hourly data from five offices in a commercial building in Alameda County between August 19, 2019, to December 31, 2021 (dates inclusive). For each office, sensors collected hourly data streams of five modalities represented in Table \ref{tab:inputvar}, which are indoor $CO_{2}$ ($CO_{2-in}$), time of the day in the hour (Hour),  number of weekdays ranging from 1 to 7 (Num-week), indoor temperature ($T_{-in}$), and outdoor temperature ($T_{-out}$) \footnote{https://bbd.labworks.org/ds/bbd/lbnlbldg59}. 

It should be noted that the number of inputs (i.e., modalities) in the dataset was initially 9, including indoor and outdoor $CO_{2}$, indoor-related humidity (RH), indoor formaldehyde, and indoor Total Volatile Organic Compounds (TVOC). These inputs were reduced to the five above by considering the contributing factors of each pollutant discovered in Mohammadshirazi et al. \cite{mohammadshirazi2022predicting}. Through the application of combinatory pattern recognition, five random sampling repetitions, and four different Ratios of Testing-Training (RTT) values of 5\%, 10\%, 15\%, and 20\%, the authors were able to select the five input variables that were most effective for indoor pollutants approximation. The total number of outputs is 5, each corresponding to a different office in the building; output is the pollutant concentration of focus (here $CO_{2}$) certain hours ahead (e.g., 1, 2, 3, or 6 hours).

\begin{table}[]
\caption{Inputs used for each predicted variable after first down-selection phase}
\label{tab:inputvar}
\begin{tabular}{c|l}
\hline
Indoor concentration predicted for the next hour   & $CO_{2-in}$ \\ \hline
\multirow{5}{*}{Current hour input variables used} & $CO_{2-in}$ \\
                                                   & $Hour$      \\
                                                   & $T_{-in}$   \\
                                                   & $T_{-out}$  \\
                                                   & $Num-week$  
\end{tabular}
\end{table}

\begin{figure}[h!]
  \includegraphics[scale=0.16]{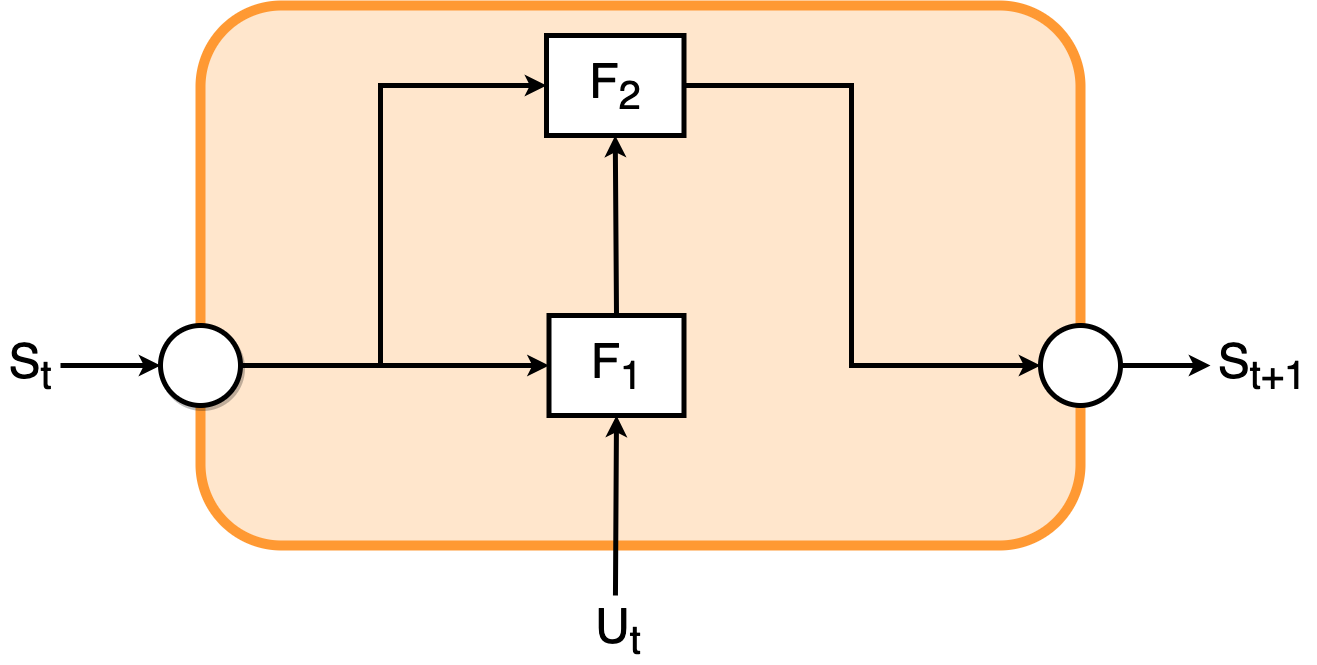}
  \caption{State-Space Recurrent Neural Network (SS-RNN).}
  \label{fig:SSRNN}
\end{figure}

\begin{figure}[h!]
  \includegraphics[scale=0.16]{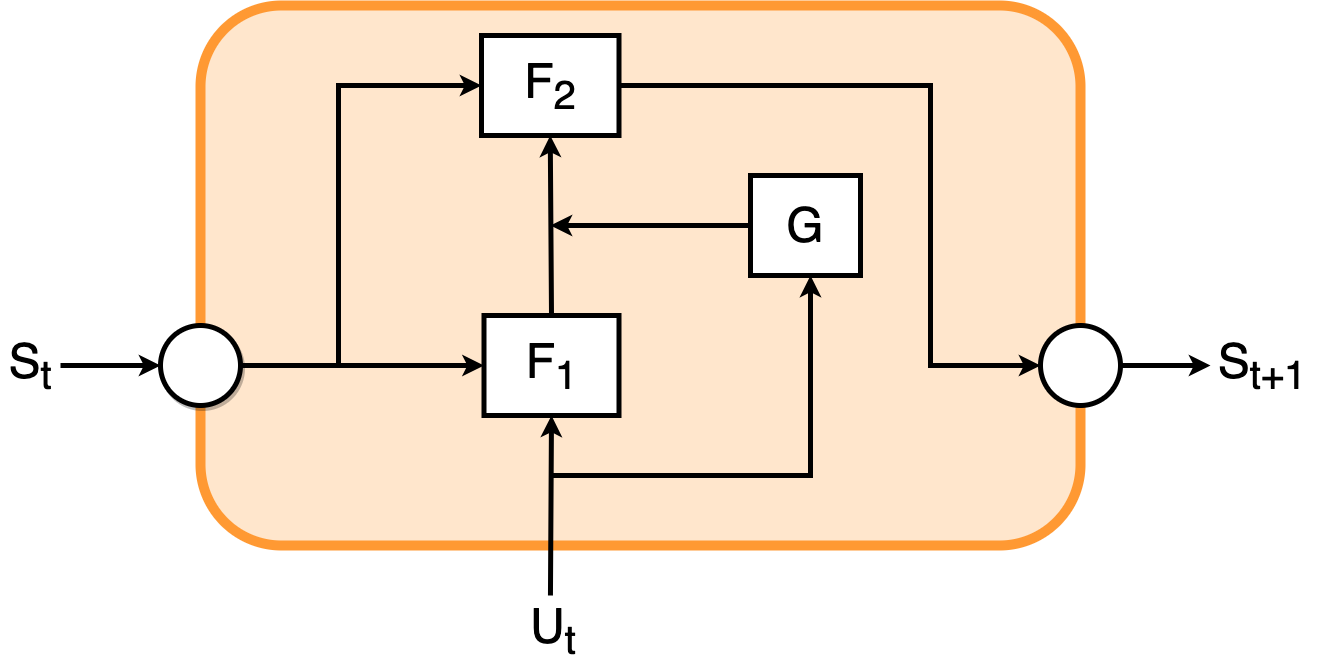}
  \caption{State-Space Gated Recurrent Unit (SS-GRU).}
  \label{fig:SSGRU}
\end{figure}

\subsection{The Six Proposed Physics-Based Models}

In this section, the six State-Space physic-based proposed models for IAQ approximation are detailed. 

\subsubsection{State-Space Recurrent Neural Networks (SS-RNN)}

In State-Space RNN (SS-RNN), the state of the network at each time step is represented by a set of continuous variables, and the dynamics of the system are represented by a set of differential equations. This allows the SS-RNN to capture more complex temporal dependencies than traditional RNNs, capturing the physic-based nonlinear patterns and domain knowledge embedded in the IAQ multi-modality temporal dataset. Besides, in contrast to traditional RNNs and LSTMs, SS-RNNs take into account the input vector's influence on the hidden states and the predicted outcome. This allows the SS-RNN to effectively capture phenomena such as mass capacitance in the input vector and diffusion or infiltration/ventilation in the predicted target. Furthermore, since SS-RNNs are explicitly designed to model the dynamics of building air quality, the only adjustable parameter is the number of hidden states determined by the geometry and pollutants in the building's air. Ultimately, in SS-RNN, the state (aka physics) variables are updated by the state equation defined by differential equations rather than conventional ML weights. It should be noted that the state equation is a function of the current state and the input. However, the state transition function is defined by the state equation, the current state, and the input.

The SS-RNN model, as detailed in Figure \ref{fig:SSRNN}, operates by taking two inputs during each temporal interval: the preceding state vector, designated as $S_{t}$, and the input vector at time $t$, denoted as $U_{t}$. An update within each cell of the model ensues for the state vector, thereby allowing for the integration of past knowledge into the current process. The transformation of the preceding state vector, $S_{t}$, along with the present input vector, $U_{t}$, is achieved through the application of a nonlinear function, $F_{1}$, which determines the change in the state vector, denoted as $dS_{t+1}$. Upon completion of this stage, the modified state vector $dS_{t+1}$ and the state vector $S_{t}$ are processed through a subsequent function, $F_{2}$, leading to the subsequent updating of $S_{t+1}$. In terms of physical interpretation, the SS-RNN model encapsulates the concept of mass capacitance as follows:

\begin{equation}
\frac{\partial }{\partial t}CO_{2\_in} = \frac {\dot m }{\rho V} (CO_{2\_out} - CO_{2\_in}) + \frac {M_{t} }{\rho}
\end{equation}

where $\dot{m}$ represents the mass flow rate related to infiltration and ventilation processes, $t$ symbolizes time, $\rho$ denotes the air density, and $V$ stands for air volume. Further, $M_t$ corresponds to the generation mass of air pollution, while $CO_{2_out}$ signifies the concentration of outdoor carbon dioxide. Each of these variables plays crucial roles in the modeling and analysis of the given physical problem, and their interplay contributes to the effectiveness of the SS-RNN model.

\begin{equation}
\underbrace{\frac{\partial }{\partial t}(CO_{2\_in})}_{dS_{t}} = A \times \underbrace{CO_{2\_in}}_{S_{t}} +\{ \underbrace {{CO_{2\_out} } + \frac {M_{t}}{\rho}}_{B {\times} {U_{t}}} \}
\end{equation}

where A and B are the state (system) matrix and input matrix, respectively. 

\begin{equation}
S_{t+1} \leftarrow  {CO_{2\_in_{t+1}}} 
\end{equation}

\begin{equation}
U_{t} \leftarrow  {CO_{2\_in_{t}}}, Hour_{t}, T\_in_{t}, T\_out_{t}, Num\_week_{t} 
\end{equation}

Based on Figure \ref{fig:SSRNN} representation, the SS-RNN ML model state space and its changes are  updated as follows:

\begin{equation}
dS_{t+1} \leftarrow F_{1}(S_{t} \times W_{dSS} + U_{t} \times W_{dSU} + b_{dS}) 
\end{equation}

\begin{equation}
S_{t+1} \leftarrow F_{2}(dS_{t+1} \times W_{SdS} + S_{t} \times W_{SS} + b_{S})
\end{equation}

In the provided equations, the terms $W_{dSS}$, $W_{dSU}$, $W_{SdS}$, and $W_{SS}$ are weight matrices that map the influence of one set of variables to another within the context of the SS-RNN model. These weights determine the level of contribution from $S_{t}$, $U_{t}$, and $dS_{t+1}$ to the new state vectors, $dS_{t+1}$ and $S_{t+1}$. The terms $b_{dS}$ and $b_{S}$ are bias vectors that contribute a constant factor to the system, allowing for a degree of flexibility and offset in the model. The activation functions $F_1$ and $F_2$ are ReLU (Rectified Linear Unit) type, allowing for the introduction of non-linearity into the model.

\subsubsection{State-Space Gated Recurrent Unit (SS-GRU)}

SS-GRU is similar to SS-RNN in the sense that they are both RNN-based models of the state-space of the physical system. However, the primary difference is in the type of gate used to update the state of the network. In SS-RNN, the gate is a nonlinear function that updates the state vector based on the input and previous state vectors. However, in SS-GRU, the gate is a GRU, a type of neural network unit that uses a gate to control the flow of information into the hidden state. The GRU acts as a decontaminator and an information transmitter from the upper layers, a popular practice observed in dense nets \cite{dey2017gate}, StyleGAN generators \cite{gal2022stylegan}, and super-resolution towers \cite{muhammad2019multi}. The use of the GRU gates in SS-GRU allows it to capture the temporal dependencies in the data more effectively and eliminates the need for a separate memory cell, which can make the model more efficient. Additionally, SS-GRU can be trained with a back-propagation through time (BPTT) algorithm, which is more efficient than traditional RNN training methods. The ML equations for SS-GRU, as depicted in Figure \ref{fig:SSGRU} are as follows:

\begin{equation}
\tilde{dS}_{t+1} \leftarrow F_{1}(S_{t} \times W_{dSS} + U_{t} \times W_{dSU} + b_{dS}) 
\end{equation}

\begin{equation}
\Gamma_{u} \leftarrow G(S_{t} \times W_{uS} + U_{t} \times W_{uU} + b_{u})
\end{equation}

\begin{equation}
dS_{t+1} \leftarrow \Gamma_{u} \times \tilde{dS}_{t+1} + (1-\Gamma_{u})S_{t}
\end{equation}

\begin{equation}
S_{t+1} \leftarrow F_{2}(dS_{t+1} \times W_{SdS} + S_{t} \times W_{SS} + b_{S})
\end{equation}

$\tilde{dS}{t+1}$ is the preliminary state vector, calculated by processing the previous state vector $S{t}$ and current input vector $U_{t}$ through the activation function $F_{1}$. $\Gamma_{u}$, determined through a sigmoid function $G$, is a gating factor that controls how much of the previous state $S_{t}$ and the preliminary state $\tilde{dS}{t+1}$ contributes to the updated state $dS{t+1}$. This gating mechanism allows the SS-GRU model to capture temporal dependencies effectively.

\subsubsection{Decomposition State-Space Recurrent Neural Networks (D-SS-RNN)}

\begin{figure}[H]
   \begin{subfigure}{1\columnwidth}
      \includegraphics[width=\linewidth]{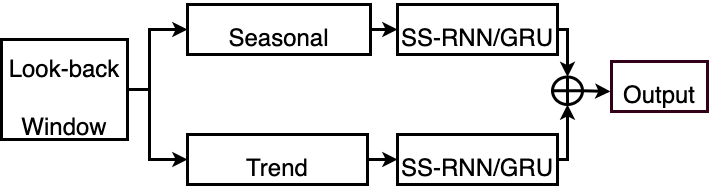}
      \caption{}
      \label{fig:DSSGRU}
   \end{subfigure}

   \begin{subfigure}{1\columnwidth}
      \includegraphics[width=\linewidth]{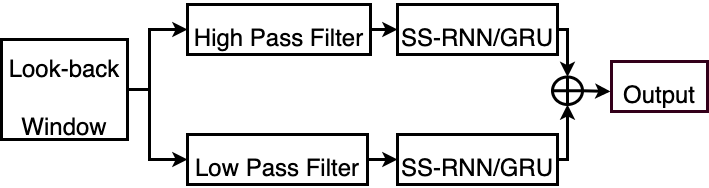}
      \caption{}
      \label{fig:graph2}
   \end{subfigure}
\caption{(a) Decomposition State-Space Recurrent Neural Network (D-SS-RNN)/ Gated Recurrent Unit (D-SS-GRU), and (b) Fourier Decomposition State-Space Recurrent Neural Network (FD-SS-RNN)/ Gated Recurrent Unit (FD-SS-GRU).}\label{fig:FDSSGRU}
\end{figure}

D-SS-RNN is a combination of SS-RNN \cite{durbin2012time} and Seasonal-Trend Decomposition \cite{wen2019robuststl}. As shown in Figure \ref{fig:DSSGRU}, the decomposition scheme builds on the idea of using moving average kernels with different kernel sizes. As depicted in Figure \ref{fig:DSSGRU}, the D-SS-RNN model combines the Seasonal-Trend Decomposition method with SS-RNN to improve its performance. Trend (moving averages) can be interpreted as a dynamic filter, as they can extract certain patterns (e.g., physics) or trends from time series data. In this way, a moving average can smooth out short-term fluctuations in the data and reveal long-term trends. Like other dynamic filters, moving averages can extract specific characteristics from a signal, such as its trend, seasonality, and cycles. The choice of window size and type of moving average will determine the information that can be extracted.

\begin{table*}[t]
    \caption{The training and testing MSEs of three state-of-the-art and six proposed models for $CO_{2}$ predicting across different dime intervals. }
    \label{tab:dssrnn}
\scalebox{1}{
\begin{tabular}{l|ll|ll|ll|ll}
\hline
\multirow{2}{*}{Model} & \multicolumn{2}{c|}{1 Hour Ahead}               & \multicolumn{2}{c|}{2 Hour Ahead}               & \multicolumn{2}{c|}{3 Hour Ahead}               & \multicolumn{2}{c}{6 Hour Ahead}               \\ \cline{2-9} 
                       & \multicolumn{1}{l|}{Training} & Testing & \multicolumn{1}{l|}{Training} & Testing & \multicolumn{1}{l|}{Training} & Testing & \multicolumn{1}{l|}{Training} & Testing \\ \hline
Transformer            & \multicolumn{1}{l|}{0.000742}     & 0.000981    & \multicolumn{1}{l|}{0.001204}     & 0.001723    & \multicolumn{1}{l|}{0.002296}     & 0.002726    & \multicolumn{1}{l|}{0.002940}      & 0.003653    \\
Informer               & \multicolumn{1}{l|}{0.000771}     & 0.001131    & \multicolumn{1}{l|}{0.001313}     & 0.001594    & \multicolumn{1}{l|}{0.002148}     & 0.002517    & \multicolumn{1}{l|}{0.003320}      & 0.004244    \\
FEDformer              & \multicolumn{1}{l|}{0.000632}     & 0.000828    & \multicolumn{1}{l|}{0.001049}     & 0.001285    & \multicolumn{1}{l|}{0.001626}     & 0.001968    & \multicolumn{1}{l|}{0.002571}     & 0.003191    \\
SS-RNN                  & \multicolumn{1}{l|}{0.000492}     & 0.000612    & \multicolumn{1}{l|}{0.000845}     & 0.001015    & \multicolumn{1}{l|}{0.001022}     & 0.001349    & \multicolumn{1}{l|}{0.002031}     & 0.002674    \\
D-SS-RNN                 & \multicolumn{1}{l|}{0.000429}     & 0.000561    & \multicolumn{1}{l|}{0.000871}      & 0.001173    & \multicolumn{1}{l|}{0.001014}     & 0.001322    & \multicolumn{1}{l|}{0.001921}     & 0.002408    \\
FD-SS-RNN                & \multicolumn{1}{l|}{0.000404}     & 0.000579    & \multicolumn{1}{l|}{0.000645}     & 0.000789    & \multicolumn{1}{l|}{0.001060}      & 0.001309    & \multicolumn{1}{l|}{0.002176}     & 0.002524    \\
SS-GRU                  & \multicolumn{1}{l|}{0.000331}     & 0.000445    & \multicolumn{1}{l|}{0.000501}     & 0.000725    & \multicolumn{1}{l|}{0.000861}      & 0.001189    & \multicolumn{1}{l|}{0.002014}     & 0.002338    \\
D-SS-GRU                 & \multicolumn{1}{l|}{\textbf{0.000329}}     & {\textbf{0.000382}}    & \multicolumn{1}{l|} {\textbf{0.000419}}     & {\textbf{0.000600}}    & \multicolumn{1}{l|} {\textbf{0.000726}}     & 0.001042    & \multicolumn{1}{l|}{0.001684}     & 0.002226    \\
FD-SS-GRU                & \multicolumn{1}{l|}{0.000368}     & 0.000443    & \multicolumn{1}{l|}{0.000470}      & 0.000637    & \multicolumn{1}{l|}{0.000785}     & {\textbf{0.000988}}    & \multicolumn{1}{l|}{\textbf{0.001667}}     & {\textbf{0.001997}}    
\end{tabular}}
\end{table*}

\begin{table}[htp]
  \caption{The six proposed and three baseline models' total number of parameters, time duration of each training epoch, and memory usage.}
  \label{tab:param}
\begin{tabular}{l|lll}
\hline
Method      & Parameter & Time   & Memory  \\ \hline
RNN         & 596.4K    & 1.7ms  & 1021MiB \\
Transformer & 13.61M    & 34.2ms & 6091MiB \\
FEDformer   & 20.68M    & 46.9ms & 4143MiB \\
Informer    & 14.39M    & 58.9ms & 3869MiB \\ \hline
SS-RNN      & 625.2K    & 2.1ms  & 1189MiB \\
SS-GRU      & 678.9K    & 2.4ms  & 1251MiB \\
D-SS-RNN    & 765.5K    & 3.0ms  & 1321MiB \\
D-SS-GRU    & 838.8K    & 3.3ms  & 1377MiB \\
FD-SS-RNN   & 872.7K    & 3.7ms  & 1491MiB \\
FD-SS-GRU   & 973.1K    & 4.1ms  & 1547MiB
\end{tabular}
\end{table}

\subsubsection{Decomposition State-Space Gated Recurrent Unit (D-SS-GRU)}

D-SS-GRU is similar to D-SS-RNN, but here GRU is replaced with an RNN component. D-SS-GRU benefits from both decomposition techniques and GRU simultaneously for pattern recognition and capturing the underlying physics and domain knowledge in the IAQ data (see Figure \ref{fig:DSSGRU}).

\subsubsection{Fourier Decomposition State-Space Recurrent Neural Networks (FD-SS-RNN)}

The SS-RNN model is generalized to decompose time series data using low-pass and high-pass filters (rather than solely focusing on seasonal and trend components-Figure \ref{fig:DSSGRU}). This approach was promising in studies such as Van et al. \cite{van2021high} and Kalauzi et al. \cite{kalauzi2023structure}. In this approach, the low-pass filter extracts the short-term physic-based trend component of the IAQ data by allowing low-frequency signals to pass through while blocking high-frequency signals. On the other hand, a high-pass filter can extract the short-term physic-based seasonal component of the IAQ data by allowing high-frequency signals to pass through while blocking low-frequency signals. Using a combination of low-pass and high-pass filters, the model discovers trends and patterns in the data that may not have been apparent when only focusing on seasonal and trend components. As such, FD-SS-RNN is proposed as a potential physics-based model for accurate IAQ approximations (Figure \ref{fig:FDSSGRU}b).

\subsubsection{Fourier Decomposition State-Space Gated Recurrent Unit (FD-SS-GRU)}

FD-SS-GRU is similar to FD-SS-RNN, but here GRU is replaced with an RNN component. D-SS-GRU benefits from both ow-pass and high-pass filtering techniques, and GRU simultaneously for pattern recognition and capturing the underlying physics and domain knowledge in the IAQ data (Figure \ref{fig:FDSSGRU}b).

\subsection{Transformer-Based Models for Comparisons}

The performance of the proposed models is compared with each other and to that of state-of-the-art non-physics-based models that are Transformers \cite{zhou2021informer, lim2021temporal}, FEDformer \cite{zerveas2021transformer}, and Informer \cite{zhou2021informer}. These baseline models were selected amongst many other (perhaps simpler) models, primarily due to the traces of elements that capture physics and domain knowledge in datasets such as IAQ. For example, in a practical order, transformer-based towers are usually enriched with multi-layer decomposition, seasonal and trend, and frequency attention layers/components, but often with complex, computationally intensive architecture.

The Transformer model, introduced by Vaswani et al. in 2017 \cite{vaswani2017attention}, is a neural network architecture used for natural language processing tasks such as machine translation, language modeling, and text summarization. It is based on the idea of attention mechanisms, which allow the model to focus selectively on certain parts of the input when making predictions. A transformer uses self-attention, which allows the model to weigh the importance of different words in the input when making predictions about a specific word. This self-attention mechanism is implemented using a multi-head attention mechanism, which allows the model to attend to different parts of the input in parallel. Additionally, the model uses a feed-forward neural network to process the input and applies layer normalization to the output of each layer to improve stability and reduce the risk of overfitting. The transformer model is trained using the standard back-propagation algorithm and is able to be parallelized easily.

The Informer model, introduced by Zhou et al. in 2021 \cite{zhou2021informer}, is a neural network architecture for text-to-speech synthesis. It is based on the transformer architecture and is trained to generate natural-sounding speech from text input. The Informer model improves upon the transformer model by incorporating a global style token and a local style token, allowing the model to generate speech with different styles, such as different speaking styles, accents, and emotions. Additionally, the Informer model uses a multi-band Mel-spectrogram as an acoustic representation of speech, which captures the characteristics of human speech more accurately than traditional spectrograms. The Informer model is trained using the standard backpropagation algorithm, and it has been demonstrated to generate speech that is more natural-sounding and expressive than previous text-to-speech models.

The FEDformer model, introduced by Zhou et al. in 2022 \cite{zhou2022fedformer}, is a neural network architecture that aims to improve fine-grained text classification by using a Federated Learning approach. The FEDformer model is based on the transformer architecture, allowing it to process variable-length text inputs efficiently. The FEDformer model aims to improve the performance of fine-grained text classification by using a Federated Learning approach, which allows the model to learn from a decentralized set of clients, each with their own unique data distribution. In this way, FEDformer can leverage the knowledge from multiple sources and overcome the problem of data insufficiency, which is common in fine-grained text classification. The model also introduced a "Federated Attention" mechanism to adapt the attention weights to the local data distribution of each client. The FEDformer model has been demonstrated to improve the performance of fine-grained text classification on several benchmark datasets.

These three transformer-based models are used as baselines for the accuracy and efficiency comparison of the six proposed models on the dataset mentioned above.

\begin{figure*}[h!]
  \includegraphics[scale=0.35]{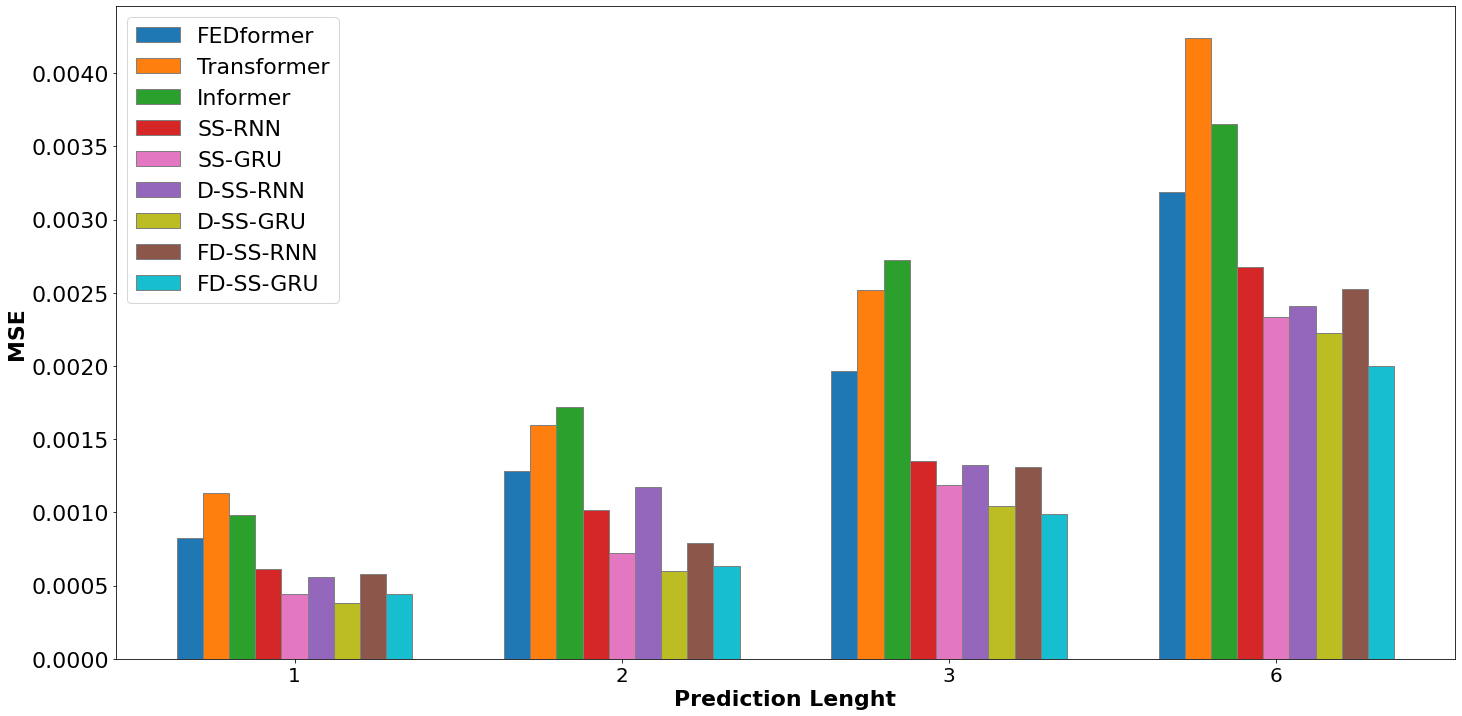}
  \caption{Methods comparison for $CO_2$ 1, 2, 3, and 6 hours ahead.}
  \label{fig:comp}
\end{figure*}

\section{Results and Analysis}
\label{sec:result}

In our experiments, the six proposed models were developed in Python 3+ environment using Tensorflow 2+ library and with one NVIDIA Volta V100 GPU accelerator (32 GB) at the Ohio Supercomputer Center (OSC) \footnote{https://www.osc.edu/}. The training stopping criteria include maximum training epochs of 600 and ten validation loss patience epochs. The training batch size was set to 128 for every proposed model. 

Table \ref{tab:5offices} provides a detailed comparison of the forecasting performances of three state-of-the-art baseline models against six proposed models, specifically in relation to $CO_{2}s$ emissions. These forecasts are conducted over several temporal spans, including 1, 2, 3, and 6-hour intervals, for each of the five office locations under study. On the other hand, Table \ref{tab:dssrnn} offers a summary representation of the same data, presenting averaged figures across all locations. It continues the comparative analysis of these three baselines and six proposed models, again focusing on $CO_{2}s$ emissions forecasting over the same time intervals.

Among all models scrutinized in this study, two proposed models D-SS-GRU and FD-SS-GRU demonstrate superior performance. They consistently outperformed other models in both training and testing phases, yielding the lowest Mean Squared Errors (MSEs).

\begin{table*}[t]
    \caption{The training and testing MSEs of three state-of-the-art and six proposed models for $CO_{2}$ predicting across different time intervals for 5 offices.}
    \label{tab:5offices}
    \centering
    \scalebox{1}{
    \begin{tabular}{l|l|ll|ll|ll|ll}
        \hline
        \multirow{2}{*}{Office} & \multirow{2}{*}{Model} & \multicolumn{2}{c|}{1 Hour Ahead} & \multicolumn{2}{c|}{2 Hour Ahead} & \multicolumn{2}{c|}{3 Hour Ahead} & \multicolumn{2}{c}{6 Hour Ahead} \\
        \cline{3-10}
                              & & \multicolumn{1}{l|}{Training} & Testing & \multicolumn{1}{l|}{Training} & Testing & \multicolumn{1}{l|}{Training} & Testing & \multicolumn{1}{l|}{Training} & Testing \\
        \hline
       \multirow{2}{*}{\#1} &Transformer & \multicolumn{1}{l|}{0.000711} & 0.000864 & \multicolumn{1}{l|}{0.001073} & 0.001952 & \multicolumn{1}{l|}{0.002519} & 0.002547 & \multicolumn{1}{l|}{0.003401} & 0.003941 \\
        &Informer    & \multicolumn{1}{l|}{0.000846} & 0.001053 & \multicolumn{1}{l|}{0.001430} & 0.001595 & \multicolumn{1}{l|}{0.002458} & 0.002895 & \multicolumn{1}{l|}{0.003671} & 0.004279 \\
       & FEDformer   & \multicolumn{1}{l|}{0.000563} & 0.000873 & \multicolumn{1}{l|}{0.001139} & 0.001217 & \multicolumn{1}{l|}{0.001650} & 0.001762 & \multicolumn{1}{l|}{0.002586} & 0.003428 \\
        &SS-RNN      & \multicolumn{1}{l|}{0.000538} & 0.000644 & \multicolumn{1}{l|}{0.000853} & 0.001015 & \multicolumn{1}{l|}{0.001125} & 0.001259 & \multicolumn{1}{l|}{0.001800} & 0.002469 \\
        &D-SS-RNN    & \multicolumn{1}{l|}{0.000429} & 0.000581 & \multicolumn{1}{l|}{0.000990} & 0.001077 & \multicolumn{1}{l|}{0.001058} & 0.001322 & \multicolumn{1}{l|}{0.002023} & 0.002221 \\
        &FD-SS-RNN   & \multicolumn{1}{l|}{0.000454} & 0.000602 & \multicolumn{1}{l|}{0.000663} & 0.000821 & \multicolumn{1}{l|}{0.001150} & 0.001498 & \multicolumn{1}{l|}{0.002431} & 0.002566 \\
        &SS-GRU      & \multicolumn{1}{l|}{0.000346} & 0.000507 & \multicolumn{1}{l|}{0.000554} & 0.000780 & \multicolumn{1}{l|}{0.000830} & 0.001325 & \multicolumn{1}{l|}{0.002113} & 0.002663 \\
       & D-SS-GRU    & \multicolumn{1}{l|}{\textbf{0.000306}} & {\textbf{0.000386}} & \multicolumn{1}{l|}{\textbf{0.000375}} & {\textbf{0.000600}} & \multicolumn{1}{l|}{\textbf{0.000695}} & {\textbf{0.000988}} & \multicolumn{1}{l|}{\textbf{0.001550}} & 0.002131 \\
        &FD-SS-GRU   & \multicolumn{1}{l|}{0.000359} & 0.000417 & \multicolumn{1}{l|}{0.000501} & 0.000610 & \multicolumn{1}{l|}{0.000909} & 0.001084 & \multicolumn{1}{l|}{0.001862} & {\textbf{0.001790}} \\
   
        \hline
        \multirow{2}{*}{\#2} &Transformer & \multicolumn{1}{l|}{0.000655} & 0.001133 & \multicolumn{1}{l|}{0.001161} & 0.001771 & \multicolumn{1}{l|}{0.002107} & 0.002473 & \multicolumn{1}{l|}{0.002603} & 0.004027 \\
        &Informer & \multicolumn{1}{l|}{0.000877} & 0.001245 & \multicolumn{1}{l|}{0.001405} & 0.001801 & \multicolumn{1}{l|}{0.002348} & 0.002731 & \multicolumn{1}{l|}{0.003223} & 0.004405 \\
        &FEDformer & \multicolumn{1}{l|}{0.000673} & 0.000849 & \multicolumn{1}{l|}{0.001122} & 0.001262 & \multicolumn{1}{l|}{0.001855} & 0.002232 & \multicolumn{1}{l|}{0.002692} & 0.003503 \\
        &SS-RNN & \multicolumn{1}{l|}{0.000488} & 0.000623 & \multicolumn{1}{l|}{0.00087} & 0.000911 & \multicolumn{1}{l|}{0.000954} & 0.001245 & \multicolumn{1}{l|}{0.002007} & 0.002922 \\
        &D-SS-RNN & \multicolumn{1}{l|}{0.000407} & 0.000564 & \multicolumn{1}{l|}{0.000809} & 0.001151 & \multicolumn{1}{l|}{0.001136} & 0.001414 & \multicolumn{1}{l|}{0.002047} & 0.002649 \\
        &FD-SS-RNN & \multicolumn{1}{l|}{0.000448} & 0.000576 & \multicolumn{1}{l|}{0.000671} & 0.00085 & \multicolumn{1}{l|}{0.000934} & 0.001193 & \multicolumn{1}{l|}{0.001919} & 0.002547 \\
        &SS-GRU & \multicolumn{1}{l|}{0.000364} & 0.000464 & \multicolumn{1}{l|}{0.000461} & 0.00083 & \multicolumn{1}{l|}{0.000827} & 0.001268 & \multicolumn{1}{l|}{0.002153} & 0.002448 \\
       & D-SS-GRU & \multicolumn{1}{l|}{\textbf{0.000301}} & {\textbf{0.000429}} & \multicolumn{1}{l|}{\textbf{0.000394}} & 0.000649 & \multicolumn{1}{l|}{\textbf{0.000732}} & {\textbf{0.000982}} & \multicolumn{1}{l|}{\textbf{0.001578}} & 0.002122 \\
        &FD-SS-GRU & \multicolumn{1}{l|}{0.000353} & 0.000438 & \multicolumn{1}{l|}{0.000446} & {\textbf{0.000576}} & \multicolumn{1}{l|}{0.000888} & 0.001116 & \multicolumn{1}{l|}{0.001605} & {\textbf{0.001791}} \\

        \hline
        \multirow{2}{*}{\#3} &Transformer   & \multicolumn{1}{l|}{0.000802} & 0.001029 & \multicolumn{1}{l|}{0.001209} & 0.001699 & \multicolumn{1}{l|}{0.002435} & 0.002736 & \multicolumn{1}{l|}{0.002912} & 0.003215 \\
        &Informer      & \multicolumn{1}{l|}{0.000753} & 0.001038 & \multicolumn{1}{l|}{0.001447} & 0.00167  & \multicolumn{1}{l|}{0.002119} & 0.002796 & \multicolumn{1}{l|}{0.003221} & 0.003849 \\
        &FEDformer     & \multicolumn{1}{l|}{0.000614} & 0.000829 & \multicolumn{1}{l|}{0.001099} & 0.001479 & \multicolumn{1}{l|}{0.001682} & 0.001874 & \multicolumn{1}{l|}{0.002626} & 0.003222 \\
        &SS-RNN        & \multicolumn{1}{l|}{0.000502} & 0.000685 & \multicolumn{1}{l|}{0.000768} & 0.00098  & \multicolumn{1}{l|}{0.000971} & 0.001546 & \multicolumn{1}{l|}{0.002249} & 0.002835 \\
        &D-SS-RNN      & \multicolumn{1}{l|}{0.000496} & 0.000628 & \multicolumn{1}{l|}{0.000883} & 0.001146 & \multicolumn{1}{l|}{0.001104} & 0.001435 & \multicolumn{1}{l|}{0.001846} & 0.002524 \\
        &FD-SS-RNN     & \multicolumn{1}{l|}{0.000364} & 0.00051  & \multicolumn{1}{l|}{0.00065}  & 0.000887 & \multicolumn{1}{l|}{0.000939} & 0.001452 & \multicolumn{1}{l|}{0.002467} & 0.002472 \\
        &SS-GRU        & \multicolumn{1}{l|}{\textbf{0.000301}} & {\textbf{0.000414}} & \multicolumn{1}{l|}{0.000446} & 0.00076  & \multicolumn{1}{l|}{0.000935} & 0.001226 & \multicolumn{1}{l|}{0.001908} & 0.002203 \\
        &D-SS-GRU      & \multicolumn{1}{l|}{0.000301} & 0.000432 & \multicolumn{1}{l|}{0.000475} & {\textbf{0.00055}}  & \multicolumn{1}{l|}{\textbf{0.000691}} & 0.001005 & \multicolumn{1}{l|}{\textbf{0.001485}} & 0.00203 \\
        &FD-SS-GRU     & \multicolumn{1}{l|}{0.000413} & 0.000469 & \multicolumn{1}{l|}{\textbf{0.000425}} & 0.000733 & \multicolumn{1}{l|}{0.000762} & {\textbf{0.000881}} & \multicolumn{1}{l|}{0.001566} & {\textbf{0.001809}} \\

        \hline
        \multirow{2}{*}{\#4} &Transformer & \multicolumn{1}{l|}{0.000714} & 0.000892 & \multicolumn{1}{l|}{0.001137} & 0.001799 & \multicolumn{1}{l|}{0.002272} & 0.003128 & \multicolumn{1}{l|}{0.002649} & 0.004124 \\
        &Informer & \multicolumn{1}{l|}{0.000684} & 0.001121 & \multicolumn{1}{l|}{0.001248} & 0.001582 & \multicolumn{1}{l|}{0.002304} & 0.002498 & \multicolumn{1}{l|}{0.003404} & 0.004078 \\
        &FEDformer & \multicolumn{1}{l|}{0.000633} & 0.000825 & \multicolumn{1}{l|}{0.001153} & 0.001309 & \multicolumn{1}{l|}{0.001481} & 0.002176 & \multicolumn{1}{l|}{0.002331} & 0.002914 \\
        &SS-RNN & \multicolumn{1}{l|}{0.00046} & 0.000695 & \multicolumn{1}{l|}{0.000791} & 0.001044 & \multicolumn{1}{l|}{0.001003} & 0.001475 & \multicolumn{1}{l|}{0.002076} & 0.002749 \\
        &D-SS-RNN & \multicolumn{1}{l|}{0.000455} & 0.000532 & \multicolumn{1}{l|}{0.000906} & 0.001324 & \multicolumn{1}{l|}{0.000923} & 0.001375 & \multicolumn{1}{l|}{0.001948} & 0.002611 \\
        &FD-SS-RNN & \multicolumn{1}{l|}{0.000426} & 0.000551 & \multicolumn{1}{l|}{0.000616} & 0.000892 & \multicolumn{1}{l|}{0.001209} & 0.001449 & \multicolumn{1}{l|}{0.00234} & 0.002234 \\
        &SS-GRU & \multicolumn{1}{l|}{0.000318} & 0.000442 & \multicolumn{1}{l|}{0.000518} & 0.000786 & \multicolumn{1}{l|}{0.00091} & 0.001355 & \multicolumn{1}{l|}{0.002225} & 0.002627 \\
        &D-SS-GRU & \multicolumn{1}{l|}{\textbf{0.000303}} & {\textbf{0.000343}} & \multicolumn{1}{l|}{0.00046} & {\textbf{0.000661}} & \multicolumn{1}{l|}{0.000809} & 0.001074 & \multicolumn{1}{l|}{0.001857} & 0.002061 \\
        &FD-SS-GRU & \multicolumn{1}{l|}{0.000362} & 0.000407 & \multicolumn{1}{l|}{\textbf{0.000449}} & 0.000731 & \multicolumn{1}{l|}{\textbf{0.000747}} & {\textbf{0.000875}} & \multicolumn{1}{l|}{\textbf{0.001637}} & {\textbf{0.002107}} \\

        \hline
        \multirow{2}{*}{\#5} &Transformer & \multicolumn{1}{l|}{0.000726} & 0.00081 & \multicolumn{1}{l|}{0.001145} & 0.001668 & \multicolumn{1}{l|}{0.002176} & 0.002166 & \multicolumn{1}{l|}{0.003253} & 0.003267 \\
        &Informer & \multicolumn{1}{l|}{0.000755} & 0.001209 & \multicolumn{1}{l|}{0.00127} & 0.001103 & \multicolumn{1}{l|}{0.001582} & 0.002187 & \multicolumn{1}{l|}{0.003517} & 0.005062 \\
        &FEDformer & \multicolumn{1}{l|}{0.000634} & 0.000703 & \multicolumn{1}{l|}{0.000727} & 0.00115 & \multicolumn{1}{l|}{0.00151} & 0.001541 & \multicolumn{1}{l|}{0.002792} & 0.003175 \\
        &SS-RNN & \multicolumn{1}{l|}{0.000469} & 0.000448 & \multicolumn{1}{l|}{0.001004} & 0.000992 & \multicolumn{1}{l|}{0.001032} & 0.001191 & \multicolumn{1}{l|}{0.001775} & 0.002243 \\
        &D-SS-RNN & \multicolumn{1}{l|}{0.000354} & 0.000463 & \multicolumn{1}{l|}{0.000954} & 0.000989 & \multicolumn{1}{l|}{0.000958} & 0.000885 & \multicolumn{1}{l|}{0.001987} & 0.002065 \\
        &FD-SS-RNN & \multicolumn{1}{l|}{0.000423} & 0.000681 & \multicolumn{1}{l|}{0.000676} & 0.000526 & \multicolumn{1}{l|}{0.001069} & 0.001161 & \multicolumn{1}{l|}{0.001828} & 0.002911 \\
        &SS-GRU & \multicolumn{1}{l|}{\textbf{0.000308}} & 0.000392 & \multicolumn{1}{l|}{0.000605} & 0.000548 & \multicolumn{1}{l|}{\textbf{0.000675}} & {\textbf{0.000805}} & \multicolumn{1}{l|}{\textbf{0.001752}} & 0.002102 \\
       &D-SS-GRU & \multicolumn{1}{l|}{0.000374} & {\textbf{0.000364}} & \multicolumn{1}{l|}{\textbf{0.00038}} & {\textbf{0.000464}} & \multicolumn{1}{l|}{0.000707} & 0.001213 & \multicolumn{1}{l|}{0.001991} & 0.002597 \\
        &FD-SS-GRU & \multicolumn{1}{l|}{0.000363} & 0.000494 & \multicolumn{1}{l|}{0.00058} & 0.000568 & \multicolumn{1}{l|}{0.000797} & 0.001017 & \multicolumn{1}{l|}{0.002009} & {\textbf{0.002076}} \\
    \end{tabular}}
\end{table*}

Table \ref{tab:param} provides a comparative illustration of the total parameters, epoch training durations for one-hour-ahead prediction, and memory usage of the six proposed models versus three established baseline models. A noteworthy observation is the relative efficiency of the SS-GRU, D-SS-GRU, and FD-SS-GRU models, exhibiting running time efficiencies of 14.25, 10.36, and 8.34 times respectively when contrasted with the Transformer model. Moreover, the computational complexity of these six models is notably lower due to the reduced parameter counts compared to the leading-edge baseline models.

The table additionally displays the inference time averages based on four runs. For instance, the inference requirement for the Informer model stands at 59ms and 3869 MiB (4 GB), whereas the FD-SS-GRU model needs only 4ms and 1547MiB (1.6 GB) for the same process. Addressing the reviewer's feedback, we intend to incorporate further information on our models' training durations. To illustrate, we will specify that the FD-SS-GRU model averaged a training time of 5.2 hours across four runs, while the Informer model needed roughly 21 hours, each using a single NVIDIA Volta V100 32 GB GPU at the OSC. Additionally, we plan to include details on time and memory complexity; for instance, the complexity for traditional transformers is $O(L^2)$; for the Informer model, it is $O(L\log(L))$, and for our six proposed models it is $O(L)$, respectively.

Predicting pollutant levels 1, 2, 3, and 6 hours in advance yielded a wide range of results depending on the method (Table \ref{tab:dssrnn} and Figure \ref{fig:comp}). Figure \ref{fig:comp} compares the performance of 9 different models, with the x-axis indicating the prediction length and the y-axis representing the MSE of the predictions. 

In terms of MSEs, D-SS-GRU and FD-SS-GRU were consistently the best models for predicting indoor pollutants (Figure \ref{fig:comp}). As Table \ref{tab:dssrnn} illustrates, the results show that D-SS-GRU performed best when predicting 1 hour ahead, with an MSE of 0.000329 for the training data and 0.000382 for the testing data. However, as the prediction interval increased to 2 hours and 3 hours, the MSE for the testing data increased to 0.000600 and 0.001042, respectively, indicating that the model's accuracy decreased as the prediction interval increased. It can be concluded that D-SS-GRU would be a model candidate for relatively short-term IAQ approximation. On the other hand, FD-SS-GRU performed better than DSSGRU when predicting further ahead, precisely at 3 hours ahead, with an MSE of 0.000988 for the testing data. This confirms that the combination of GRU and low- and high-pass filters would provide a stronger mid- to long-term IAQ approximation by capturing physics-based patterns hidden in the nonlinear IAQ temporal records. 

For 6 hours ahead, FD-SS-GRU's MSE for testing and training data were 0.001997 and 0.001667, respectively. This suggests that FD-SS-GRU is more accurate than D-SS-GRU in long-term approximation. This superiority of SS-GRU over SS-RNN can be attributed to the gating mechanism in GRUs. The gating mechanism allows the GRU to effectively control the flow of input variables ($U_t$ in figure \ref{fig:SSGRU}) and decide which portion of the input to retain and which to discard. The gating mechanism helps the GRU filter the input variables more effectively than an RNN without a gating mechanism (SS-RNN). Additionally, the gating mechanism helps the GRU better handle the vanishing gradient problem, which can be a common issue in traditional RNNs. These advantages of GRUs over RNNs can be seen in the improved performance results, as shown in table \ref{tab:dssrnn} and figure \ref{fig:comp}. Overall, models consisting of GRU showed lower MSEs than the ones with RNN.
\section{Conclusion and Future Work}
\label{sec:conlcusion}
In this study, we proposed six novel physics-based ML models for accurate indoor pollutant concentration approximations by an adroit combination of State-Space concepts in physics, Gated Recurrent Units, and Decomposition techniques. The proposed models were shown to be less complex, computationally more efficient, and more accurate than similar state-of-the-art transformer-based models that are Transformers \cite{zhou2021informer, lim2021temporal}, FEDformer \cite{zerveas2021transformer}, and Informer \cite{zhou2021informer}. The superiority of the proposed models is due to their relatively light architecture (computational efficiency) and, more importantly, their ability to capture the underlying highly nonlinear patterns embedded in the often contaminated sensor-collected IAQ temporal data.

\bibliographystyle{ACM-Reference-Format}
\bibliography{references}

\end{document}